\documentclass[10pt,twocolumn,letterpaper]{article}

\usepackage[pagenumbers]{iccv} 

\usepackage{textcomp, gensymb}
\usepackage{bm}
\usepackage{booktabs}
\definecolor{iccvblue}{rgb}{0.21,0.49,0.74}
\usepackage[breaklinks,colorlinks,allcolors=iccvblue]{hyperref}
\usepackage[accsupp]{axessibility}
\usepackage{balance}
\usepackage{tikz}

\title{Guiding Diffusion-Based Articulated Object Generation by Partial Point Cloud Alignment and Physical Plausibility Constraints}

\author{Jens U. Kreber and Joerg Stueckler\\
University of Augsburg\\
Augsburg, Germany\\
{\tt\small \{jens.kreber,joerg.stueckler\}@uni-a.de}
}

\newcommand\copyrighttext{%
\footnotesize \textcopyright~2025 IEEE. Personal use of this material is permitted.
Permission from IEEE must be obtained for all other uses, in any current or future
media, including reprinting/republishing this material for advertising or promotional
purposes, creating new collective works, for resale or redistribution to servers or
lists, or reuse of any copyrighted component of this work in other works.
}
\newcommand\copyrightnotice{%
\begin{tikzpicture}[remember picture,overlay]
\node[anchor=south,yshift=10pt] at (current page.south) {\fbox{\parbox{\dimexpr\textwidth-\fboxsep-\fboxrule\relax}{\copyrighttext}}};
\end{tikzpicture}%
}

\newcommand\acceptancetext{%
\centering
\footnotesize Accepted for publication at the IEEE/CVF International Conference on Computer Vision (ICCV), 2025}
\newcommand\acceptancenotice{%
\begin{tikzpicture}[remember picture,overlay]
\node[anchor=north,yshift=-10pt] at (current page.north)
{\parbox{\dimexpr\textwidth-\fboxsep-\fboxrule\relax}
{\acceptancetext}};
\end{tikzpicture}%
}

\begin{document}
\maketitle

\acceptancenotice
\copyrightnotice
\vspace{-\baselineskip}

\begin{abstract}
Articulated objects are an important type of interactable objects in everyday environments.
In this paper, we propose PhysNAP, a novel diffusion model-based approach for generating articulated objects that aligns them with partial point clouds and improves their physical plausibility.
The model represents part shapes by signed distance functions (SDFs).
We guide the reverse diffusion process using a point cloud alignment loss computed using the predicted SDFs.
Additionally, we impose non-penetration and mobility constraints based on the part SDFs for guiding the model to generate more physically plausible objects.
We also make our diffusion approach category-aware to further improve point cloud alignment if category information is available.
We evaluate the generative ability and constraint consistency of samples generated with PhysNAP using the PartNet-Mobility dataset.
We also compare it with an unguided baseline diffusion model and demonstrate that PhysNAP can improve constraint consistency and provides a tradeoff with generative ability.
\end{abstract}

\section{Introduction}
Many man-made objects in everyday environments such as drawers, appliances, or laptops are articulated.
Generating and reconstructing digital twins of such articulated objects is an important problem in virtual reality applications and robotics.

\begin{figure}[tb]
    \centering
    \includegraphics[width=\linewidth]{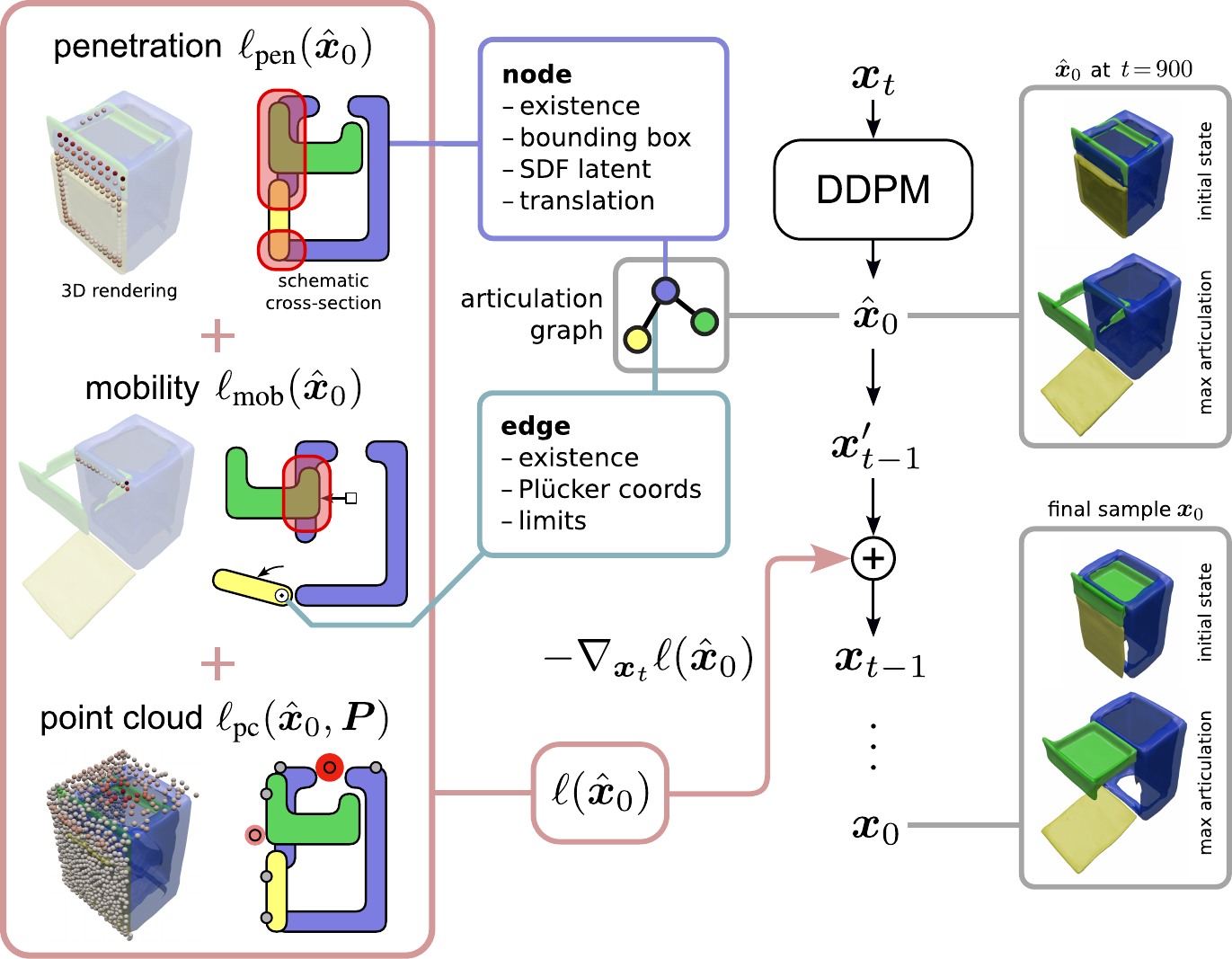}
    \caption{Method overview of PhysNAP: We propose to guide a pretrained diffusion model (NAP~\cite{lei2023_nap}) for articulated object generation with point cloud alignment and physical plausibility losses. The latter measure part penetration in the initial state (penetration) and in varying articulation states (mobility).
    All generated properties can potentially be adjusted to follow the losses, including node and edge existence, shape and articulation joint properties.
    We also augment NAP to condition on the object category.
    }
    \label{fig:teaser}
\end{figure}

In this paper, we propose PhysNAP, a novel diffusion model-based approach that generates articulated objects that align with partial point cloud views by loss-guided sampling.
We include guidance loss terms which improve physical plausibility of the generated objects by reducing part penetrations in the initial state (zero articulation) and in varying articulation states.
Additionally, we condition the diffusion model on object categories which also improves point cloud alignment, since with category information the model is more likely to generate well-matching objects.
An overview of our method is shown in \Cref{fig:teaser}.
Our approach extends NAP~\cite{lei2023_nap}, an unconditional and unguided diffusion model which generates articulation graphs, part shapes, and joint parameters.
We propose novel approaches to calculate point distance and shape penetration based on the predicted graph representation and signed distance function (SDF) part shapes.
The losses are differentiable by all generated properties, which allows for multiple ways of minimizing them, such as adjusting the relative location of objects or altering their shape or articulation properties.
To the best of our knowledge, PhysNAP is the first category-aware diffusion model which can generate articulated objects with point cloud alignment and physical plausibility guidance without prior knowledge of the articulation graph structure, i.e. number of existing parts and connectivity by joints.

We evaluate PhysNAP on the PartNet-Mobility dataset~\cite{chang2015shapenet, Mo_2019_partnet, Xiang_2020_SAPIEN} and demonstrate that it improves point cloud alignment and physical plausibility over an unconditional, unguided baseline approach.
We provide ablations to assess the benefit of our design choices and analyze tradeoffs between guidance loss compliance, category-awareness, and generative ability.

In summary, we make the following contributions:
\begin{itemize}
    \item We propose PhysNAP, a novel diffusion model-based approach that generates articulated objects by extending NAP to condition on object categories and training-free guidance by 3D measurements and physical plausibility.
    \item We guide the reverse diffusion process with point cloud alignment and physical plausibility constraints that are determined from the predicted articulation graph and SDF part shapes. Physical plausibility is measured by penetration of parts in the initial and varying articulation states.
    \item We assess the performance of PhysNAP in terms of generative ability, point cloud alignment, and physical plausibility. Our approach improves on the latter two while providing a tradeoff with generative ability.
\end{itemize}

\section{Related work}
\label{sec:rel-work}

Generating articulated objects and reconstructing them from measurements such as images and point clouds is an important step for creating digital twins for virtual reality applications and robotics. 
The approaches can be classified by output representations for articulation and shapes, inference methods,
and conditioning information.

\textbf{Output representation.}
The methods in~\cite{jiang2022_ditto,liu2023_paris,mu2021_asdf} learn part shapes and parameters of a single joint from point cloud or image observations.
Recently, methods have been proposed such as~\cite{gao2024_meshart,su2024_artformer,chen2024_urdformer} which predict multi-joint articulation graphs with joint parameters and part shapes.
ArticulateAnything~\cite{le2025_articulateanything} predicts this information in URDF format.
Real2Code~\cite{mandi2025_real2code} predicts code descriptions of articulated objects.
While~\cite{liu2023_paris} represents part shapes as neural radiance fields (NeRFs), the approaches in~\cite{mu2021_asdf,su2024_artformer,lei2023_nap,leboutet2024_midgard,luo2025_physpart}  yield part shapes as latent embeddings of signed distance functions (SDFs) and Real2Code~\cite{mandi2025_real2code} infers part shapes as occupancy grids from which meshes can be extracted.
The methods in~\cite{gao2024_meshart,chen2024_urdformer,le2025_articulateanything,liu2024_cage,liu2025_singapo} output meshes for the part shapes.
Our approach predicts part shapes as latent embedding vectors which are decoded into SDFs as in NAP~\cite{lei2023_nap}.
By this, shape can be continuously varied by the latent vector during the reverse diffusion process.
PhysPart~\cite{luo2025_physpart} uses a latent 3D feature grid for representing SDF shape, while we use more compact latent feature vectors.

\textbf{Inference method.}
Several methods use supervised learning to regress articulation models of objects. 
Ditto~\cite{jiang2022_ditto} uses a PointNet-based network architecture.
PARIS~\cite{liu2023_paris} uses deep learning to optimize NeRF representations and joint parameters of the articulated object from multiple images.
Both methods only support articulated objects consisting of two parts and a single joint.
The approaches in~\cite{gao2024_meshart,su2024_artformer} use transformer-based architectures to regress articulation graphs with multiple joints.
URDFormer~\cite{chen2024_urdformer} transforms rendered images from simulated objects to realistic images using a text-conditional image diffusion model.
This way, paired image and articulated object descriptions (URDFs with joint parameters) are obtained which are used for supervised training of deep regression models.
Some recent works~\cite{le2025_articulateanything,mandi2025_real2code} use vision and language foundation models (LLMs, VLMs) to regress articulated object models from images.
Differently, our approach leverages a diffusion model as generative prior of articulated objects.
Inference from point measurements is performed by training-free guidance.
A-SDF~\cite{mu2021_asdf} learns a generative auto-decoder model which parameterizes the articulated object with a single joint angle and the shape latent codes for two parts.
Inference is performed by optimizing shape latent and joint angle to minimize the SDF of an input point cloud. 
Several previous methods have been proposed that learn diffusion models of articulated objects such as~\cite{lei2023_nap,liu2024_cage,liu2025_singapo,leboutet2024_midgard,luo2025_physpart}.
Our approach belongs to the latter category.
We extend NAP~\cite{lei2023_nap} to guide the reverse diffusion process by point cloud alignment and physical plausibility constraints.

\textbf{Conditioning information.} 
Some methods such as~\cite{gao2024_meshart,lei2023_nap} generate articulated objects without conditioning information.
Ditto~\cite{jiang2022_ditto} regresses the articulation model from two point cloud observations.
PARIS~\cite{liu2023_paris} requires multiple image views as input for NeRF optimization.
ArtFormer~\cite{su2024_artformer} conditions the transformer model on text instructions.
A-SDF~\cite{mu2021_asdf} uses an input point cloud for optimization-based inference.
The approaches in ~\cite{chen2024_urdformer,le2025_articulateanything,mandi2025_real2code} take single images of the articulated object as input.
CAGE~\cite{liu2024_cage} conditions a diffusion model on object category and, differently to our approach, requires the articulation graph structure as input.
In SINGAPO~\cite{liu2025_singapo} the model is extended to  condition on RGB image features. 
The connectivity graph is regressed using a vision-language foundation model (VLM).
Like our approach, the diffusion model approach MIDGaRD~\cite{leboutet2024_midgard} does not require the articulation graph as input.
Also differently to our approach, it generates articulation graph structure and shapes in subsequent stages which prevents loss guidance for the graph through the predicted shape.

Closely related to our approach is PhysPart~\cite{luo2025_physpart}.
The approach assumes that a full point cloud of an incomplete object base is available as input and generates physically plausible part completions and joint parameters by conditioning the diffusion model on point cloud features.
Similar to our approach, PhysPart also applies loss guidance using part penetration and mobility constraints determined from SDF shape representations of the parts.
Differently, our approach is more general, as we do not assume the articulation graph structure given and use partial point clouds for training-free guidance. PhysPart requires a full point cloud of the object base for conditioning.
Moreover, PhysPart only penalizes collisions between the base object and a single generated part connected to it (SDF).
Our approach determines penetration losses between all parts and mobility losses between connected parts.

\section{Method}
We propose PhysNAP, a novel approach for generating physically plausible articulated objects given a partial point cloud view of the object.
We extend NAP~\cite{lei2023_nap} which uses a Denoising Diffusion Probabilistic Model (DDPM)~\cite{Ho.etal2020DenoisingDiffusionProbabilisticModels} by conditioning it on object class information.
In addition, the reverse diffusion process is guided to align the parts of the articulated object with the partial input point cloud.
We also guide the reverse diffusion to avoid penetrations between parts in the initial state of the object and when being actuated.
In the following we will detail our method.

\subsection{Denoising Diffusion Probabilistic Models}

Denoising diffusion probabilistic models (DDPM,~\cite{Ho.etal2020DenoisingDiffusionProbabilisticModels}) model the diffusion of a data sample~$\bm{x}_0$ by a latent variable~$\bm{x}_t$ in successive time steps~$t$.
For the forward diffusion process, DDPM assumes that
\begin{equation}
q(\bm{x}_t \mid \bm{x}_0) = \mathcal{N}\left(\bm{x}_t; \sqrt{\bar{\alpha}_t} \bm{x}_0, (1 - \bar{\alpha}_t) \bm{I}\right)
\end{equation}
is a Gaussian distribution conditional on the data sample where $\bar{\alpha}_t$ is a factor as defined in~\cite{Ho.etal2020DenoisingDiffusionProbabilisticModels}.
This allows to write  $\bm{x}_t(\bm{x}_0, \bm{\epsilon}) = \sqrt{\bar{\alpha}_t} \bm{x}_0 + \sqrt{1 - \bar{\alpha}_t} \bm{\epsilon}$ for $\bm{\epsilon} \sim \mathcal{N}(\bm{0},\bm{I})$ in terms of $\bm{x}_0$ and $\bm{\epsilon}=\bm{\epsilon}(\bm{x}_t, \bm{x}_0)$~\cite{Ho.etal2020DenoisingDiffusionProbabilisticModels}.
By successive diffusion, the latent variable tends to $\bm{x}_T \sim \mathcal{N}(\bm{0}, \bm{I})$ for $T \rightarrow \infty$.

It can be shown (see~\cite{Ho.etal2020DenoisingDiffusionProbabilisticModels}) that this diffusion process can be reverted by iteratively sampling a latent variable for the previous time step according to
\begin{equation}\label{eq:ddpm_sampling}
    \bm{x}_{t-1} = \frac{1}{\sqrt{\alpha_t}} \left( \bm{x}_t - \frac{1 - \alpha_t}{\sqrt{1 - \bar{\alpha}_t}} \epsilon_{\bm{\theta}}(\bm{x}_t, t) \right) + \sigma_t \bm{z}
\end{equation}
where $\bm{z} \sim \mathcal{N}(\bm{0},\bm{I})$, $\sigma_t$ is a noise scale parameter, and $\alpha_t$ is a factor as defined in~\cite{Ho.etal2020DenoisingDiffusionProbabilisticModels}.
To this end, the noise~$\bm{\epsilon}(\bm{x}_t, \bm{x}_0)$ of the forward diffusion process is approximated by a neural network $\bm{\epsilon}_{\bm{\theta}}(\bm{x}_t,t)$ with parameters~${\bm{\theta}}$ which is trained to predict the forward diffusion process noise from~$\bm{x}_t$ and~$t$. 
The reverse diffusion process is initialized with a sample $\bm{x}_T \sim \mathcal{N}(\bm{0}, \bm{I})$ at time step~$T$.
Note that $\bm{\epsilon}_{\bm{\theta}}(\bm{x}_t,t)$ is closely related to the score of $p_t(\bm{x}_t) = \int q(\bm{x}_t \mid \bm{x}_0) p(\bm{x}_0) d \bm{x}_0 $ and it holds $\nabla_{\bm{x}_t} \log p_t(\bm{x}_t) \approx  - \frac{1}{\sqrt{1 - \bar{\alpha}_t}} \bm{\epsilon}_{\bm{\theta}}(\bm{x}_t, t)$ 
since
$\nabla_{\bm{x}_t} \log q(\bm{x}_t \mid \bm{x}_0) = - \frac{1}{\sqrt{1 - \bar{\alpha}_t}} \left( \bm{x}_t - \sqrt{\bar{\alpha}_t} \bm{x}_0  \right) 
= - \frac{1}{\sqrt{1 - \bar{\alpha}_t}} \bm{\epsilon}(\bm{x}_t, \bm{x}_0).
$

\subsection{Loss-Guided Diffusion}
Now to make the denoising process conditional on measurements~$\bm{y}$, one replaces $\nabla_{\bm{x}_t} \log p_t(\bm{x}_t)$ by 
\begin{equation}
\nabla_{\bm{x}_t} \log p_t(\bm{x}_t \mid \bm{y}) = \nabla_{\bm{x}_t} \log p_t(\bm{x}_t) + \nabla_{\bm{x}_t} \log p_t(\bm{y} \mid \bm{x}_t)    
\end{equation}
as a result of Bayes rule~\cite{DiffusionPosteriorSampling}.
However, $p_t(\bm{y} \mid \bm{x}_t)$ is intractable to compute~\cite{DiffusionPosteriorSampling}, while we can determine $p_t(\bm{y} \mid \bm{x}_t) \approx p_t(\bm{y} \mid \hat{\bm{x}}_0)$.
We approximate the posterior mean $\hat{\bm{x}}_0 := \mathbb{E}[\bm{x}_0\mid \bm{x}_t]$ by $\hat{\bm{x}}_0 \approx \frac{1}{\sqrt{\bar{\alpha}_t}} \left( \bm{x}_t - \sqrt{1 - \bar{\alpha}_t} \bm{\epsilon}_{\bm{\theta}}(\bm{x}_t,t) \right)$.

Following loss-guided diffusion~\cite{LossGuidedDiffusion}, we calculate $p(\bm{y} | \hat{\bm{x}}_0) = \frac{1}{Z} \exp(-\ell_{\bm{y}}(\hat{\bm{x}}_0))$, where $\ell_{\bm{y}}(\hat{\bm{x}}_0)$ is a loss function for~$\hat{\bm{x}}_0$ given measurement~$\bm{y}$ and $Z$ is the partition function which marginalizes over~$\hat{\bm{x}}_0$ and, hence, is independent of the specific~$\hat{\bm{x}}_0$.
This allows to approximate
$
\nabla_{\bm{x}_t} \log p_t(\bm{y} \mid \bm{x}_t) \approx \nabla_{\bm{x}_t} \log p_t(\bm{y} \mid \hat{\bm{x}}_0)
= -\nabla_{\bm{x}_t} \ell_{\bm{y}}(\hat{\bm{x}}_0).
$
To implement the guidance, we follow DPS \cite{DiffusionPosteriorSampling} and add  $-\nabla_{\bm{x}_t} \ell_{\bm{y}}(\hat{\bm{x}}_0)$ to $\bm{x}_{t-1}$ in the denoising step.

\subsection{Guided Articulation Graph Diffusion} \label{sec:guide_arti}
We base our approach on NAP~\cite{lei2023_nap} which generates articulation graphs with a maximum number of~$K$ nodes ($K=8$ in our experiments).
Each node represents a part of the articulated object.
The edges in the graph represent potential articulations between pairs of parts.
The attributes of a node are an existence indicator $o_i \in \{0,1\}$, a pose $\bm{T}_{gi} \in SE(3)$ for the part's initial configuration in the global frame, bounding box edge lengths $\bm{b}_i \in \mathbb{R}^3$, and a shape latent code $\bm{s}_i \in \mathbb{R}^F$ with $F=128$ in our experiments.
A pretrained neural signed distance function (SDF) for parts with latent shape embedding~\cite{park2019_deepsdf,lei2023_nap} is used for representing shapes.
For the diffusion model, the node attributes are represented as real vectors $\bm{v}_i \in \mathbb{R}^{D_v}$ with existence indicator $o_i$ represented continuously as $\tilde{o}_i \in \mathbb{R}$.
Note that NAP assumes that for each object, an initial state with zero articulation exists in which all parts are in a canonical orientation throughout the dataset (for example, all drawers facing the same direction).
Therefore, the initial rotation of parts is assumed zero and the model only predicts 3D translations.

Edge attributes are existence indicator $c_{i,j} \in \{-1, 0, 1\}$ that also controls the direction of the edge, 6-dimensional Pl\"ucker coordinates $\bm{p}_{i,j} = (\bm{l}_{i,j},\bm{m}_{i,j})$ with $\bm{l}_{i,j} \in \mathbb{S}^2, \bm{m}_{i,j} \in \mathbb{R}^3$ that describe the articulation and joint state ranges $\bm{\rho}_{i,j} \in \mathbb{R}^{2\times2}$.
The edge attributes are represented as $\bm{e}_{i,j} \in \mathbb{R}^{D_e}$.
The edge existence is represented as a real 3D vector $\tilde{\bm{c}}_{i,j}$ which is set to the one-hot representation of $c_{i,j}$ to embed the edge.
The Plücker coordinates and joint state ranges are embedded as real vectors.

Since the chirality of the edges is modeled by an attribute, the graph consists of~$K$ nodes and~$K (K-1) / 2$ edges which can be represented as a real vector.
The diffusion model learns to predict noise on this representation using an attentional graph neural network (AGNN).
Generated samples of the graph representation are projected back to a manifold where the graph is properly defined.
This includes projecting the values of the embedded Plücker coordinates in $\mathbb{R}^6$ to its constraints.
The existence values for nodes are thresholded to determine which nodes should actually exist and the edge existence values are used to extract the minimum spanning tree.
Meshes are extracted from the part SDFs using marching cubes~\cite{marchingcubes}.
NAP centers and rescales the extracted meshes to match the statistics of the training set. 
We do not perform this post-processing step, as it would require differentiable mesh extraction and lookups of SDF values at shifted locations.
Lastly, NAP supports retrieving the most similar mesh for the predicted SDF shape in the training set.

\subsection{Category-Aware Generation}
We augment NAP to additionally condition on specific object categories for the case that such a category is assumed known.
We add a learnable category embedding to the noise prediction network architecture, which is added to the node and edge embeddings in the AGNN.
We then retrain NAP, providing it with the ground-truth categories of the object it is denoising.

\subsection{Guidance}
We design a loss function for guiding the NAP reverse diffusion process to minimize inter-part penetrations in the initial pose and when moving joints, as well as to align the generated object shape with a partial point cloud view.
In the following, we will detail the individual loss terms for inter-part penetrations ($\ell_\text{pen}(\hat{\bm{x}}_0)$), mobility ($\ell_\text{mob}(\hat{\bm{x}}_0)$), and point cloud alignment ($\ell_\text{pc}(\hat{\bm{x}}_0, \bm{P})$) for point cloud~$\bm{P}$.
The loss terms are combined by
\begin{equation}
\ell_{\bm{P}}(\hat{\bm{x}}_0) = w_\text{pc} \ell_\text{pc}(\hat{\bm{x}}_0, \bm{P}) + w_\text{pen} \ell_\text{pen}(\hat{\bm{x}}_0) + w_\text{mob} \ell_\text{mob}(\hat{\bm{x}}_0),
\end{equation}
where $w_\text{pc}, w_\text{pen}, w_\text{mob}$ are weight factors which we determine in a parameter study (see Sec.~\ref{sec:experiments}).

\subsubsection{Point Cloud Alignment Loss}
We measure the alignment of the predicted articulated object in the initial state to a partial point cloud  $\bm{P} \in \mathbb{R}^{n_p \times 3}$.
We evaluate the distance of the points to a part shape using the predicted SDF representations.
We determine the SDF value of point~$\bm{P}_j$ for part~$i$ by transforming the point into the local coordinate frame of part~$i$ using the predicted translation of node~$\bm{t}_i$ and querying the neural SDF at the transformed point.
The distance is scaled by the extents of the bounding box~$\bm{b}_i$ of the part.
If the transformed point does not lie within the bounding box, we approximate the SDF by the distance to the closest point on the bounding box plus the SDF value at this closest point.
The probabilistic point measurement model for part $i$ is a Gaussian in the signed distance with zero mean and a fixed variance (subsumed in the weight~$w_\text{pc}$).

Since the association of the points to parts is unknown, we estimate the association using the measurement model similar to expectation-maximization.
We compute a soft correspondence likelihood of the~$j$-th point to each part~$i$ based on the respective distance and existence:
\begin{equation}
    \alpha_{i,j} = \frac{\exp\left(-\tau \  d(\bm{P}_j,i)^2 \, /\, (\tilde{o}_i + \epsilon)\right)}{\sum_k \exp\left(-\tau \  d(\bm{P}_j,k)^2 \, /\, (\tilde{o}_k + \epsilon)\right)},
\end{equation}
where~$d(\bm{P}_j,i)$ is the signed distance of point~$\bm{P}_j$ to part~$i$,~$\epsilon$ is a small value for numerical stability and~$\tau$ is an inverse temperature parameter.
Note that this formulation makes the loss differentiable w.r.t. to the existence of nodes, so theoretically, a non-existing part could be introduced to match the points more closely.
In our experiments, we use a relatively low temperature of $\tau=1000$.
Finally, we compute $\ell_\text{pc}(\hat{\bm{x}}_0, \bm{P}) = \sum_j \sum_i \alpha_{i,j} d(\bm{P}_j,i)^2$.

\begin{figure*}[tb]
  \centering
  \begin{subfigure}{0.335\linewidth}
    \includegraphics[width=\linewidth]{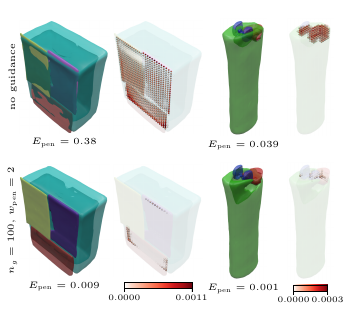}
    \caption{Example of penetration loss guidance:
    Initial state penetrations between all parts are reduced.
    \\
    }
    \label{fig:pen_qualitative}
  \end{subfigure}
  \hfill
  \begin{subfigure}{0.65\linewidth}
    \includegraphics[width=\linewidth]{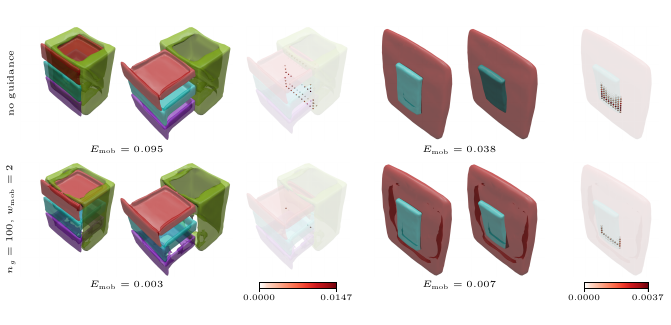}
    \caption{Example of mobility loss guidance:
    Penetrations between direct neighbors at randomly sampled articulation states are reduced.
    Left/middle: Minimal/maximal articulation state. Right: SDF penetration error at sample points in the articulation state with highest total error.
    }
    \label{fig:mob_qualitative}
  \end{subfigure}
  \caption{Visual comparison of samples from unguided and our loss-guided generation for the category-unaware model.
  Meshes are color-coded by parts.
  Plots with colorbar show the SDF penetration error at sample points, encoded by opacity/color.
  }
  \label{fig:physics_guidance}
\end{figure*}

\subsubsection{Penetration Loss}
We determine a penetration loss by measuring penetrations between all possible pairs of parts.
First, we calculate the intersection volume of the part bounding boxes using the predicted bounding box extents~$\bm{b}_i$ and poses~$\bm{T}_{gi}$.
We sample a 3D grid of points in the intersection cuboid $\bm{b}_{\text{inter}}$ and evaluate the penetration using the predicted SDF shapes.

We impose two requirements on the 3D grid sample points:
The number of samples should be limited and ideally constant to bound the computational effort.
We aim at~$N^*=1000$ samples in our experiments.
Also, we aim at sampling points with similar density in each dimension.

We determine the intersection volume~$V=\prod_{k=1}^{3} \bm{b}_{\text{inter},k}$ and~$l_V = \sqrt[3]{V}$ as the edge length of a cube with identical volume.
The number of points per dimension is~$n_k = \lfloor \max(\sqrt[3]{N^*}\,\bm{b}_{\text{inter},k} / l_V, 1)\rfloor$ and the total number of points~$N = \prod_{k=1}^3 n_k$.
Without the clamping and rounding, the number of points would be exactly~$N^*$, but could result in less than one point per dimension.
With the proposed approach, the~$N$ can also be higher than~$N^*$ for thin shapes.

For each sample $\bm{q}_j$ in our grid, we evaluate the SDF values $d(\bm{q}_j, i)$ and $d(\bm{q}_j, i')$ for the pair of parts~$i$ and~$i'$ as described in the previous section.
We then calculate the SDF penetration error as in~\cite{strecke2024_physically,luo2025_physpart}
\begin{equation}
    \psi(\bm{q}_j, i, i') = \frac{1}{2} \min \left( 0, - (d(\bm{q}_j, i) + d(\bm{q}_j, i'))  \right)^2.
\end{equation}
See \Cref{fig:pen_qualitative} for an example of the effect of guidance with penetration loss.

The penetration loss becomes
\begin{align}
\ell_\text{pen}(\hat{\bm{x}}_0) = V_\text{el} \sum_i \sum_{i'\neq i} \sum_{j=1}^N \psi(\bm{q}_j, i, i') \, \tilde{o}_i \, \tilde{o}_{i'}
\end{align}
where we scale the errors by the volume element size $V_\text{el} = 10^5 \prod_{k=1}^3 \bm{b}_{\text{inter},k} / n_k$ and the continuous existence values $\tilde{o}_i, \tilde{o}_{i'}$.
Thereby, theoretically, intersections can be resolved by removing a part.
Note that due to the overall scaling, the intersection errors are relatively small in comparison to the errors in the other loss terms.
Thus, we scale the volume element size by~$10^5$.

\subsubsection{Mobility Loss}
For the mobility loss, we measure penetration between pairs of parts in varying articulated joint states similar to~\cite{luo2025_physpart}. 
Predicting the articulation structure requires to calculate a minimum spanning tree which avoids kinematic loops.
Since this is computationally involved and not differentiable, this processing step is not suitable for our loss guidance during the reverse diffusion.
We therefore restrict the penetration evaluation to parts connected by an edge.
While this will not capture all possible intersections between parts in the predicted object, it will measure penetrations between many important parts which are neighbors in the graph such as doors or drawers and the object main body.
Differently to our method,~\cite{luo2025_physpart} only considers penetrations between the object base and articulated parts.

We only consider edges between parts $i, i'$ for which the existence value of one of the directions is above a threshold of $0.1$.
We project the Plücker coordinates onto valid values as in~\cite{lei2023_nap}.
We sample two articulation states with screw angle~$\gamma$ and joint extension~$d$ within the predicted joint limits.
The coordinate transformation between the parts is 
$\bm{T}_{ii'} = \bm{T}_{\text{screw}}(\bm{l},\bm{m}, \gamma, d) \ \bm{T}_{gi}^{-1} \ \bm{T}_{gi'}$
where $(\bm{l},\bm{m})$ are the Plücker coordinates and $\bm{T}_{\text{screw}}(\bm{l},\bm{m}, \gamma, d)$ is the transformation matrix for the screw parameters and Plücker coordinates.

We transform the bounding box of part~$i'$ into frame~$i$ and retrieve the circumscribing axis-aligned bounding box in frame~$i$.
Then we proceed as for the penetration loss by summing SDF penetration errors of points on a grid and scaling by the respective continuous edge existence value in~$\tilde{\bm{c}}_{i,j}$.
The mobility loss~$\ell_\text{mob}(\hat{\bm{x}}_0)$ is set to two times the resulting value to account for both edge directions to balance it with the penetration loss.

\section{Experiments}
\label{sec:experiments}

We address the following questions in our experiments:
(1) Can loss guidance improve point cloud alignment and physical plausibility for generated objects?
(2) Is there a tradeoff between achieving the guidance targets and generative ability?
(3) Does conditioning on the category improve results in guidance targets over the unconditional case?

\subsection{Experiment Setup}
We compare the performance of PhysNAP to ablations and to unmodified NAP~\cite{lei2023_nap} which is our direct baseline.
We do not compare to methods such as \cite{liu2024_cage,luo2025_physpart} as they require the articulation graph structure given.
We also cannot compare to \cite{leboutet2024_midgard,luo2025_physpart} as code and dataset variants are not publicly available.

Like NAP, we use 1000 diffusion steps in total for the forward and reverse diffusion processes.
For training our category-aware diffusion model, we use the same hyperparameters and training modalities (objective and dataset, except for the additional category annotations) as NAP.
NAP is trained on the PartNet-Mobility dataset~\cite{chang2015shapenet, Mo_2019_partnet, Xiang_2020_SAPIEN}.
We use their training, validation, and test splits.
We generate a dataset of observed point clouds for validation and testing on the respective splits of the PartNet-Mobility dataset.
For the test set, we sample 30 object models at random without replacement.
Additionally, we sample a random camera pose for each object.
The elevation ranges from $30\degree$ to $80\degree$ and the azimuth from $120\degree$ to $240\degree$, ensuring that objects are well observed from at least two sides.
For the camera distance, we use minimum and maximum ranges from 1 to 6.
The scaling of meshes is unchanged from PartNet-Mobility.
We reject a sampled camera pose if the bounding box in pixels spans less than 50\% horizontally and vertically.
Poses are also rejected if the object cuts the edges of the image.
We randomly sample $M=1000$ points within the object segmentation mask based on the rendering pipeline of GAPartNet~\cite{GAPartNet}.
We assume knowledge of the object pose and transform the observed point cloud into object coordinates.

For guided generative sampling, we add our guidance terms for the last~$n_g$ steps. 
We determine~$n_g$ in a hyperparameter study (see Sec.~\ref{sec:hp_search}).

\subsubsection{Evaluation Metrics}
\textbf{Guidance Metrics.}
To measure the alignment of a generated object sample to the guidance point cloud, we compute the squared SDF value of each part with existence indicator value $> 0.5$ at each point.
We take the minimum over these parts at each point, yielding the error of the closest respective part, and determine the mean squared error
    $E_\text{pc} = \frac{1}{M} \sum_{j=1}^{M} \min_{i:\tilde{o}_i > 0.5} d(\bm{p}_j, i)^2$.
We also extract a mesh from the part SDFs using marching cubes~\cite{marchingcubes} and compute the mean distance~$D_\text{pc}$ of the points to the mesh.

To measure penetration and mobility, we measure part penetrations as for our guidance losses.
We do not weight for node or edge existence, but instead only evaluate penetrations between parts with existence larger than 0.5.
For the mobility evaluation, we also do not sample random articulation states, but compute the loss for 5 equally-spaced articulation states between the minimum and maximum joint limits.
We call the respective error metrics~$E_\text{pen}$ and~$E_\text{mob}$.

\textbf{Generative Metrics.}
We adopt the generative metrics minimum matching distance (MMD) and 1-nearest neighbor accuracy (1-NNA) used in NAP~\cite{lei2023_nap}.
Details on the base metrics can be found in~\cite{yangPointFlow3DPoint2019}.
NAP adapts the metrics for the articulated case by searching the minimum of each metric over sampled articulation states.
We use these variants of the metrics.
To compute the base metrics in a particular articulated state, the Chamfer distance between 2048 points sampled randomly on the generated and extracted meshes is determined.
Since our focus lies on the meshes extracted from the SDFs, we report metrics on those meshes.
Further implementation details on the metrics can be found in the supplementary material.

\begin{figure}[tb]
  \centering
  \includegraphics{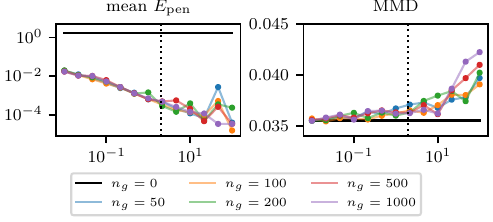}
  \caption{Hyperparameter sweep over guidance steps~$n_g$ and guidance weight $w_\text{pen}$ (x-axis) for generation with the penetration guidance loss  only, without category conditioning. Black line: NAP baseline. We choose $n_g=500$ and $w_\text{pen,base}=2$ (dashed line).}
  \label{fig:pen_sweep}
\end{figure}

\subsubsection{Hyperparameter Study} \label{sec:hp_search}
We perform a hyperparameter study to find an empirical setting of the number of guidance steps~$n_g$ and the weights~$w_*$ for the guidance losses.
We conduct a parameter sweep for each of the guidance losses independently.
In \Cref{fig:pen_sweep}, we show results for the penetration guidance loss for which we compare with the unmodified, unguided NAP baseline.
Similar to NAP, we generate a number of samples equal to the size of the validation split to calculate the generative metrics for this evaluation.
We observe that with stronger guidance, the guidance metric decreases, but the generative metric gets worse.
Additional plots are shown in the supplementary material.
We find that hyperparameter search results for $n_g=500$ and $n_g=1000$ are very similar, especially for point cloud guidance.
Presumably, in early steps, the latent variable describes coarse properties of articulation structure and shape and can still adapt to the guidance losses.
Due to the lower computational effort, we choose~$n_g = 500$ for all of our models and select the base weights~$w_\text{pc,base}=45, w_\text{pen,base}=2, w_\text{mob,base}=2$.
When combining several loss terms, we divide the base weights by the number of used terms to determine the weights ~$w_\text{pc}, w_\text{pen}, w_\text{mob}$.

\subsection{Results}
\begin{table}[tb]
\centering
\footnotesize
\setlength{\tabcolsep}{3pt}
\begin{tabular}{llcccccc}
\toprule
cat & variant & $E_\text{pc} $ & $D_\text{pc}$ & $E_\text{pen} $ & $E_\text{mob}$ & MMD & 1-NNA\\
\midrule
no & pc+pen+mob & 0.0024 & 0.0705 & \textbf{0.0000} & \textbf{0.0003} & 0.1435 & 0.9547 \\
no & pc+pen & 0.0014 & 0.0616 & 0.0001 & 0.0004 & 0.1523 & 0.9580 \\
no & pc+mob & 0.0014 & 0.0603 & 0.0002 & 0.0004 & 0.1525 & 0.9612 \\
no & pc & \textbf{0.0006} & \textbf{0.0435} & 0.0018 & 0.0031 & 0.1540 & 0.9666 \\
no & uncond \cite{lei2023_nap} & 0.0564 & 0.2063 & 0.0035 & 0.0033 & \textbf{0.0915} & \textbf{0.9440} \\
\midrule
yes & pc+pen+mob & 0.0012 & 0.0483 & \textbf{0.0000} & \textbf{0.0003} & 0.1970 & 0.9774 \\
yes & pc+pen & 0.0008 & 0.0396 & 0.0001 & 0.0005 & 0.1995 & 0.9774 \\
yes & pc+mob &0.0008 & 0.0388 & 0.0003 & 0.0006 & 0.2096 & 0.9784 \\
yes & pc & \textbf{0.0004} & \textbf{0.0301} & 0.0079 & 0.0059 & 0.2162 & 0.9817 \\
yes & uncond &0.0076 & 0.0974 & 0.0019 & 0.0027 & \textbf{0.1687} & \textbf{0.9709} \\
\bottomrule
\end{tabular}
\caption{Results of PhysNAP (pc+pen+mob) and ablations (pc: point cloud loss guidance, pen: penetration loss guidance, mob: mobility loss guidance, uncond: no guidance, cat: category conditioning). 
Lower is better for all metrics. 
Adding pc improves point cloud alignment compared to the baseline.
Adding pen+mob reduces point cloud alignment, but improves penetration and mobility losses, constituting a tradeoff.
The category-aware model generally yields better point cloud-aligned objects, but at a loss of diversity (higher 1-NNA).
Best results marked as bold.
}
\label{tab:main_results}
\end{table}

\textbf{Quantitative Results.}
We compare the results of PhysNAP with ablations in \Cref{tab:main_results}.
All guidance metrics report the median over 450 generated samples. 
The generative metrics are calculated for 15 samples per point cloud and the median is taken over 30 point clouds. 
As reference, we also evaluate an unconditional baseline (corresponding to NAP if not category-aware) that has no guidance term.
Compared to it, all variants of PhysNAP with point cloud loss guidance (pc) achieve better point cloud alignment and all variants with penetration (pen) and/or mobility loss guidance (mob) achieve better physical plausibility measures.
Addressing question (1), overall we observe, that adding the guidance terms reduces the respective errors.

We further see that solely using the pc yields the lowest~$E_\text{pc}$ 
and~$D_\text{pc}$.
Adding penetration and/or mobility losses yields worse alignment to points, but lower penetration and mobility errors and also slightly better generative metrics.
However, compared to the baseline, all variants suffer a loss in generative metrics.
Note that this can be expected since the variety of generated objects is restricted to better correspond to the point cloud.
Concluding and addressing question (2), there is indeed a tradeoff, both between achieving different guidance targets and between a single guidance target and generative performance as measured by MMD and 1-NNA.
The category-aware model variants outperform their unaware counterparts in point cloud alignment and perform similarly with respect to physical plausibility.
They do, however, have worse generative metrics.
The higher 1-NNA value is expected, since the category implies a loss in diversity of generated samples.
So regarding question (3), conditioning on a category improves point cloud alignment but at a loss of diversity.

With all of our proposed losses included, the generation process takes slightly less than 2 minutes per sample on average on an Nvidia A40.
Further run time evaluation can be found in the supplementary material.

\begin{figure*}[tb]
    \centering
    \includegraphics[width=\linewidth]{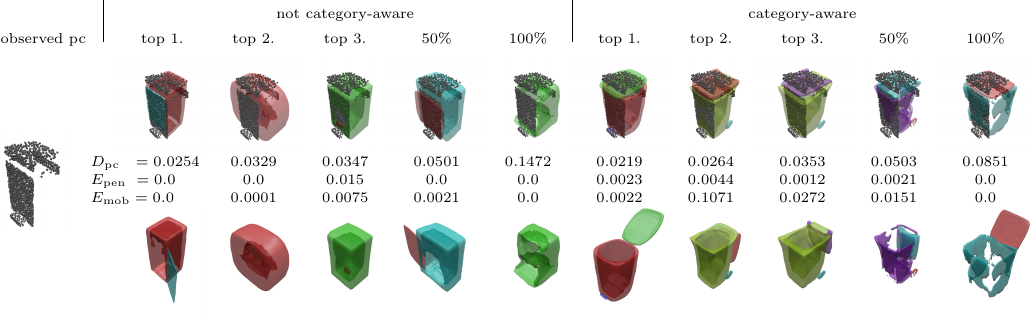}
    \noindent\rule{\textwidth}{0.5pt}
    \includegraphics[width=\linewidth]{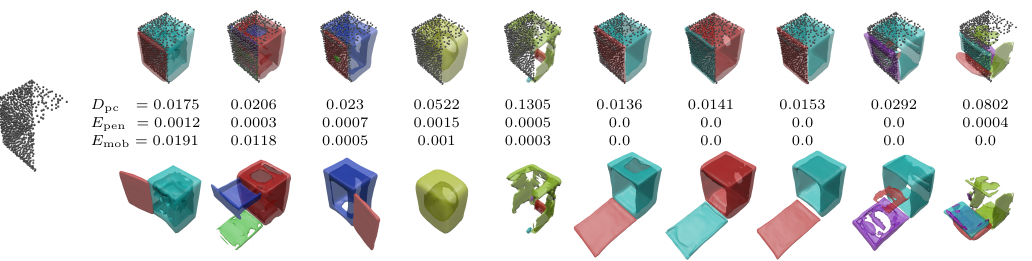}
    \noindent\rule{\textwidth}{0.5pt}
    \includegraphics[width=\linewidth]{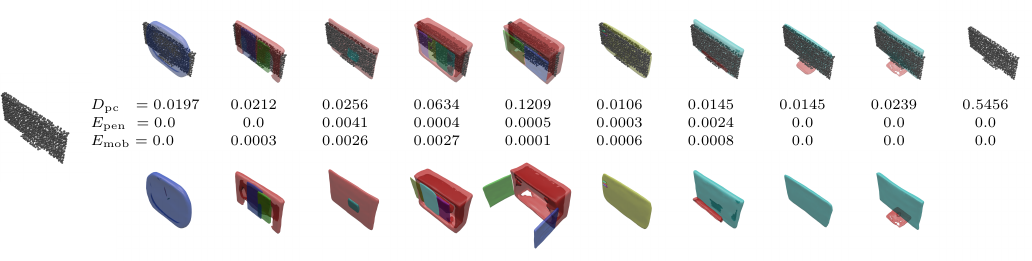}
    \noindent\rule{\textwidth}{0.5pt}
    \includegraphics[width=\linewidth]{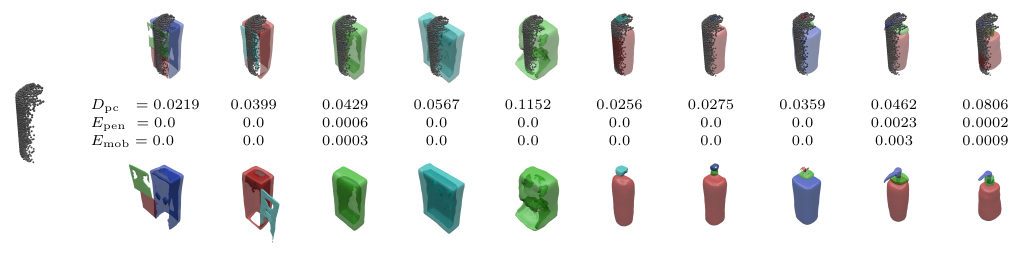}
    \caption{Articulated objects generated for the observed point clouds on the left. We sort the samples by $D_\text{pc}$ and report the 3 best as well as 50\% and 100\% (worst) quantiles.
    $D_\text{pc}$ is indicative of quality: The top results mostly have clear identifiable shapes and articulations in contrast to the worst samples.
    The worst sample for point cloud~3 from the category-aware model failed mesh extraction (no point with negative SDF has been sampled).
    A limitation of the category-unaware model can be seen especially for point cloud~4, where many of the generated shapes are wider than the point cloud.
    The category-aware model improves on that.
    }
    \label{fig:qualitative}
\end{figure*}

\textbf{Qualitative Results.}
We also evaluate PhysNAP qualitatively.
In \Cref{fig:qualitative}, we show examples of articulated objects generated from the model for given point clouds.
We sort the samples by~$D_\text{pc}$ and report the 3 best as well as 50\% and 100\% (worst) quantiles.
We can observe that~$D_\text{pc}$ coincides with intuitive quality:
The best 3 results mostly have clear identifiable shapes and articulations.
The shape of the worst cases can hardly be identified for most of them and the articulation does not appear meaningful.
The model without category awareness generates different classes of objects to match the observed shape.
The category-aware model consequently exhibits less diversity, but is able to match the point cloud better in most cases.
This is especially well visible for the last row example, where the category-unaware model generates many shapes that are wider than the point cloud, while the category-aware model matches it visibly better.
In the supplementary material, we also show examples of generated objects with retrieved part meshes.

To obtain a visual interpretation of the effect of the penetration and mobility losses, we choose two settings respectively from our hyperparameter study with only $n_g =100$ steps.
Incorporating guidance that late in the reverse diffusion process has the effect that the generated sample shape stays similar to the sample shape without guidance, if the reverse diffusion starts from same initial value and uses the same random seed for the noise $\bm{z}$.
In \Cref{fig:pen_qualitative}, we compare two unguided and penalty loss-guided samples.
Similarly, in \Cref{fig:mob_qualitative}, we compare two unguided and mobility loss-guided samples.
We can see that in both cases part penetrations are reduced for the guided samples.

\section{Conclusions}

In this paper, we propose PhysNAP, a novel generative modeling approach for articulated objects that aligns the generated objects with partial point cloud views and improves their physical plausibility.
We measure physical plausibility in terms of penetration between parts in the initial state of the object and in varying articulation states.
We make the diffusion model category-aware which further improves point cloud alignment for settings in which the category is known.
PhysNAP inherits the property of its baseline NAP that the diffusion model does not require prior knowledge of the articulation graph structure.

We demonstrate on the PartNet-Mobility dataset that PhysNAP can improve point cloud alignment and physical plausibility for generated objects.
We also provide ablations to assess the impact of our design choices and analyze the tradeoffs for our loss guidance terms and category-awareness between guidance metrics and generative ability.

We regard our approach as an important step toward perception of articulated objects for AR and robotics.
Generated objects could be used in physics simulations, which would require disabling remaining collisions between parts.
Directly generating objects without any collisions is a potential topic for future research.
Currently, our approach assumes that the sensor pose w.r.t. the canonical articulated object frame is known. 
This could be alleviated in future work, for instance, by devising a learning-based approach that predicts the relative pose of the articulated object from measurements.
Also, the quality of the generated part shapes is limited by the SDF decoder, which could be addressed in future work.
Methods such as DDIM could be explored to improve the run time of our approach.
We also plan to incorporate image information into PhysNAP to improve object reconstruction.

\section*{Acknowledgments}
This work has been supported by Hightech Agenda Bayern and Deutsche Forschungsgemeinschaft (DFG) project no. 466606396 (STU 771/1-1).

\balance

{
    \small
    \bibliographystyle{ieeenat_fullname}
    \bibliography{main}

\begin{thebibliography}{24}
\providecommand{\natexlab}[1]{#1}
\providecommand{\url}[1]{\texttt{#1}}
\expandafter\ifx\csname urlstyle\endcsname\relax
  \providecommand{\doi}[1]{doi: #1}\else
  \providecommand{\doi}{doi: \begingroup \urlstyle{rm}\Url}\fi

\bibitem[Chang et~al.(2015)Chang, Funkhouser, Guibas, Hanrahan, Huang, Li, Savarese, Savva, Song, Su, Xiao, Yi, and Yu]{chang2015shapenet}
Angel~X. Chang, Thomas~A. Funkhouser, Leonidas~J. Guibas, Pat Hanrahan, Qi{-}Xing Huang, Zimo Li, Silvio Savarese, Manolis Savva, Shuran Song, Hao Su, Jianxiong Xiao, Li Yi, and Fisher Yu.
\newblock Shape{N}et: An information-rich {3D} model repository.
\newblock \emph{CoRR}, abs/1512.03012, 2015.

\bibitem[Chen et~al.(2024)Chen, Walsman, Memmel, Mo, Fang, Fox, and Gupta]{chen2024_urdformer}
Qiuyu Chen, Aaron Walsman, Marius Memmel, Kaichun Mo, Alex Fang, Dieter Fox, and Abhishek Gupta.
\newblock {URDFormer}: {A} pipeline for constructing articulated simulation environments from real-world images.
\newblock In \emph{Proc. of Robotics: Science and Systems (RSS)}, 2024.

\bibitem[Chung et~al.(2023)Chung, Kim, McCann, Klasky, and Ye]{DiffusionPosteriorSampling}
Hyungjin Chung, Jeongsol Kim, Michael~Thompson McCann, Marc~Louis Klasky, and Jong~Chul Ye.
\newblock Diffusion posterior sampling for general noisy inverse problems.
\newblock In \emph{Proc. of Int. Conf. on Learning Representations (ICLR)}, 2023.

\bibitem[Gao et~al.(2025)Gao, Siddiqui, Li, and Dai]{gao2024_meshart}
Daoyi Gao, Yawar Siddiqui, Lei Li, and Angela Dai.
\newblock {MeshArt}: Generating articulated meshes with structure-guided transformers.
\newblock In \emph{Proc. of IEEE/CVF Conf. on Computer Vision and Pattern Recognition (CVPR)}, 2025.

\bibitem[Geng et~al.(2023)Geng, Xu, Zhao, Xu, Yi, Huang, and Wang]{GAPartNet}
Haoran Geng, Helin Xu, Chengyang Zhao, Chao Xu, Li Yi, Siyuan Huang, and He Wang.
\newblock {{GAPartNet}}: {{Cross-Category Domain-Generalizable Object Perception}} and {{Manipulation}} via {{Generalizable}} and {{Actionable Parts}}.
\newblock In \emph{Proc. of {{IEEE}}/{{CVF Conf.}} on {{Computer Vision}} and {{Pattern Recognition}} ({{CVPR}})}, 2023.

\bibitem[Ho et~al.(2020)Ho, Jain, and Abbeel]{Ho.etal2020DenoisingDiffusionProbabilisticModels}
Jonathan Ho, Ajay Jain, and Pieter Abbeel.
\newblock Denoising diffusion probabilistic models.
\newblock In \emph{Advances in Neural Information Processing Systems (NeurIPS)}, 2020.

\bibitem[Jiang et~al.(2022)Jiang, Hsu, and Zhu]{jiang2022_ditto}
Zhenyu Jiang, Cheng{-}Chun Hsu, and Yuke Zhu.
\newblock Ditto: Building digital twins of articulated objects from interaction.
\newblock In \emph{Proc. of {IEEE/CVF} Conf. on Computer Vision and Pattern Recognition (CVPR)}. {IEEE}, 2022.

\bibitem[Le et~al.(2025)Le, Xie, Liang, Wang, Yang, Ma, Vedder, Krishna, Jayaraman, and Eaton]{le2025_articulateanything}
Long Le, Jason Xie, William Liang, Hung{-}Ju Wang, Yue Yang, Yecheng~Jason Ma, Kyle Vedder, Arjun Krishna, Dinesh Jayaraman, and Eric Eaton.
\newblock {Articulate-Anything}: Automatic modeling of articulated objects via a vision-language foundation model.
\newblock In \emph{Proc. of Int. Conf. on Learning Representations (ICLR)}, 2025.

\bibitem[Leboutet et~al.(2024)Leboutet, Wiedemann, Cai, Paulitsch, and Yuan]{leboutet2024_midgard}
Quentin Leboutet, Nina Wiedemann, Zhipeng Cai, Michael Paulitsch, and Kai Yuan.
\newblock {MIDGArD}: Modular interpretable diffusion over graphs for articulated designs.
\newblock In \emph{Advances in Neural Information Processing Systems (NeurIPS)}, 2024.

\bibitem[Lei et~al.(2023)Lei, Deng, Shen, Guibas, and Daniilidis]{lei2023_nap}
Jiahui Lei, Congyue Deng, William~B. Shen, Leonidas~J. Guibas, and Kostas Daniilidis.
\newblock {NAP:} {Neural} {3D} articulated object prior.
\newblock In \emph{Advances in Neural Information Processing Systems (NeurIPS)}, 2023.

\bibitem[Liu et~al.(2023)Liu, Mahdavi{-}Amiri, and Savva]{liu2023_paris}
Jiayi Liu, Ali Mahdavi{-}Amiri, and Manolis Savva.
\newblock {PARIS:} part-level reconstruction and motion analysis for articulated objects.
\newblock In \emph{Proc. of {IEEE/CVF} Int. Conf. on Computer Vision (ICCV)}, 2023.

\bibitem[Liu et~al.(2024)Liu, Tam, Mahdavi{-}Amiri, and Savva]{liu2024_cage}
Jiayi Liu, Hou In~Ivan Tam, Ali Mahdavi{-}Amiri, and Manolis Savva.
\newblock {CAGE:} controllable articulation generation.
\newblock In \emph{Proc. of {IEEE/CVF} Conf. on Computer Vision and Pattern Recognition (CVPR)}, 2024.

\bibitem[Liu et~al.(2025)Liu, Iliash, Chang, Savva, and Mahdavi{-}Amiri]{liu2025_singapo}
Jiayi Liu, Denys Iliash, Angel~X. Chang, Manolis Savva, and Ali Mahdavi{-}Amiri.
\newblock {SINGAPO:} single image controlled generation of articulated parts in objects.
\newblock In \emph{Proc. of Int. Conf. on Learning Representations (ICLR)}, 2025.

\bibitem[Lorensen and Cline(1987)]{marchingcubes}
William~E. Lorensen and Harvey~E. Cline.
\newblock Marching cubes: {A} high resolution {3D} surface construction algorithm.
\newblock In \emph{Proc. of 14th Annual Conf. on Computer Graphics and Interactive Techniques, {SIGGRAPH}}. {ACM}, 1987.

\bibitem[Luo et~al.(2025)Luo, Geng, Deng, Li, Wang, Jia, Guibas, and Huang]{luo2025_physpart}
Rundong Luo, Haoran Geng, Congyue Deng, Puhao Li, Zan Wang, Baoxiong Jia, Leonidas~J. Guibas, and Siyuan Huang.
\newblock {PhysPart}: Physically plausible part completion for interactable objects.
\newblock In \emph{Proc. of IEEE Int. Conf. on Robotics and Automation (ICRA)}, 2025.
\newblock To appear.

\bibitem[Mandi et~al.(2025)Mandi, Weng, Bauer, and Song]{mandi2025_real2code}
Zhao Mandi, Yijia Weng, Dominik Bauer, and Shuran Song.
\newblock {Real2Code}: Reconstruct articulated objects via code generation.
\newblock In \emph{Proc. of Int. Conf. on Learning Representations (ICLR)}, 2025.

\bibitem[Mo et~al.(2019)Mo, Zhu, Chang, Yi, Tripathi, Guibas, and Su]{Mo_2019_partnet}
Kaichun Mo, Shilin Zhu, Angel~X. Chang, Li Yi, Subarna Tripathi, Leonidas~J. Guibas, and Hao Su.
\newblock {PartNet}: A large-scale benchmark for fine-grained and hierarchical part-level {3D} object understanding.
\newblock In \emph{Proc. of IEEE/CVF Conf. on Computer Vision and Pattern Recognition (CVPR)}, 2019.

\bibitem[Mu et~al.(2021)Mu, Qiu, Kortylewski, Yuille, Vasconcelos, and Wang]{mu2021_asdf}
Jiteng Mu, Weichao Qiu, Adam Kortylewski, Alan~L. Yuille, Nuno Vasconcelos, and Xiaolong Wang.
\newblock {A-SDF:} learning disentangled signed distance functions for articulated shape representation.
\newblock In \emph{Proc. of {IEEE/CVF} Int. Conf. on Computer Vision (ICCV)}, 2021.

\bibitem[Park et~al.(2019)Park, Florence, Straub, Newcombe, and Lovegrove]{park2019_deepsdf}
Jeong~Joon Park, Peter~R. Florence, Julian Straub, Richard~A. Newcombe, and Steven Lovegrove.
\newblock {DeepSDF}: Learning continuous signed distance functions for shape representation.
\newblock In \emph{Proc. of {IEEE/CVF} Conf. on Computer Vision and Pattern Recognition (CVPR)}, 2019.

\bibitem[Song et~al.(2023)Song, Zhang, Yin, Mardani, Liu, Kautz, Chen, and Vahdat]{LossGuidedDiffusion}
Jiaming Song, Qinsheng Zhang, Hongxu Yin, Morteza Mardani, Ming-Yu Liu, Jan Kautz, Yongxin Chen, and Arash Vahdat.
\newblock Loss-guided diffusion models for plug-and-play controllable generation.
\newblock In \emph{Proc. of 40th {{Int. Conf.}} on {{Machine Learning}} (ICML)}, 2023.

\bibitem[Strecke and Stueckler(2024)]{strecke2024_physically}
Michael Strecke and Joerg Stueckler.
\newblock Physically plausible object pose refinement in cluttered scenes.
\newblock In \emph{Proc. of German Conf. on Pattern Recognition (GCPR)}, 2024.

\bibitem[Su et~al.(2025)Su, Feng, Li, Song, He, Ren, and Xu]{su2024_artformer}
Jiayi Su, Youhe Feng, Zheng Li, Jinhua Song, Yangfan He, Botao Ren, and Botian Xu.
\newblock {ArtFormer}: Controllable generation of diverse {3D} articulated objects.
\newblock In \emph{Proc. of {IEEE/CVF} Conf. on Computer Vision and Pattern Recognition (CVPR)}, 2025.

\bibitem[Xiang et~al.(2020)Xiang, Qin, Mo, Xia, Zhu, Liu, Liu, Jiang, Yuan, Wang, Yi, Chang, Guibas, and Su]{Xiang_2020_SAPIEN}
Fanbo Xiang, Yuzhe Qin, Kaichun Mo, Yikuan Xia, Hao Zhu, Fangchen Liu, Minghua Liu, Hanxiao Jiang, Yifu Yuan, He Wang, Li Yi, Angel~X. Chang, Leonidas~J. Guibas, and Hao Su.
\newblock {SAPIEN}: A simulated part-based interactive environment.
\newblock In \emph{Proc. of IEEE/CVF Conf. on Computer Vision and Pattern Recognition (CVPR)}, 2020.

\bibitem[Yang et~al.(2019)Yang, Huang, Hao, Liu, Belongie, and Hariharan]{yangPointFlow3DPoint2019}
Guandao Yang, Xun Huang, Zekun Hao, Ming{-}Yu Liu, Serge~J. Belongie, and Bharath Hariharan.
\newblock {PointFlow}: {3D} point cloud generation with continuous normalizing flows.
\newblock In \emph{Proc. of {IEEE/CVF} Int. Conf. on Computer Vision (ICCV)}, 2019.

\end{thebibliography}
}

\newpage
\nobalance
\appendix
\twocolumn[
    \begin{center}
      \Large\bfseries Supplementary Material
      \vspace{3ex}
    \end{center}
]

\section{Introduction}

In this supplementary material we provide additional details and results for our approach.
We first provide details on the modifications of the open-source implementation of NAP used in our experiments.
Then, we detail results of our hyperparameter study.
Finally, we provide a table of retrieved meshes as comparison to the meshes extracted from SDFs in our method.

\section{Modifications of open-source implementation of NAP}
In the open-source implementation of NAP provided by the authors~\footnote{\url{https://github.com/JiahuiLei/NAP}}, we changed the edge feature embedding for the graph network to be consistent with the described method according to our understanding.
Additionally, we changed the order of sample and reference set in the calculation of the generative metrics which is more consistent with the original formulation of the metrics for our evaluation.
For reference, we compare our computed metrics on the baseline (unguided) method against the values obtained with the open-source implementation of NAP (both with the modified edge embedding) and also provide the numbers as reported in the NAP paper.
We evaluate with 5 different sample sets from different random seeds and compute the mean and standard deviation (in parentheses).
See \Cref{tab:metrics_unguided}.
It can be seen that with the modified feature embedding but old metric implementation (NAP metric), results differ only slightly from those reported in the original paper (from NAP).
With the adapted metrics (ours), the deviation from the original results further increases.
Additionally to the metrics MMD and 1-NNA, we report the COV (coverage) metric as in the NAP paper.
As for the other metrics, details on the base metrics can be found in~\cite{yangPointFlow3DPoint2019}.

\begin{table*}[tb]
  \centering
    \begin{tabular}{cccc}
    \toprule
    &MMD $\downarrow$ (ours) & MMD $\downarrow$ (NAP metric \cite{lei2023_nap}) & MMD $\downarrow$ (from NAP \cite{lei2023_nap})\\
    \midrule
    SDF & 0.0284 (0.0004) & 0.0272 (0.0012) & 0.0268 \\
    retrieval & 0.0265 (0.0008) & 0.0226 (0.0013) & 0.0215 \\
    \midrule
 &COV $\uparrow$ (ours) & COV $\uparrow$ (NAP metric \cite{lei2023_nap}) & COV $\uparrow$ (from NAP \cite{lei2023_nap}) \\
 \midrule
 SDF& 0.4833 (0.0109) & 0.4976 (0.0097) & 0.4944\\
 retrieval & 0.4806 (0.0131) & 0.5301 (0.0143) & 0.5234\\
 \midrule
        &1-NNA $\downarrow$ (ours) & 1-NNA $\downarrow$ (NAP metric \cite{lei2023_nap}) & 1-NNA $\downarrow$ (from NAP \cite{lei2023_nap}) \\
        \midrule
       SDF & 0.6477 (0.0204) & 0.5615 (0.0057) & 0.5690\\
       retrieval & 0.5739 (0.0227) & 0.5367 (0.0102) & 0.5412\\
       \bottomrule
  \end{tabular}
  \caption{Comparison of our modified implementation of NAP and generative metric computation with the open-source implementations of NAP and its metrics. Ours: modified edge feature embedding; NAP metric: metric from NAP open-source implementation with our modified edge feature embedding; from NAP: results reported in original paper. Values are given as mean (standard deviation).}
  \label{tab:metrics_unguided}
\end{table*}

\section{Hyperparameter Study}

\begin{figure*}[tb]
  \centering
  \includegraphics{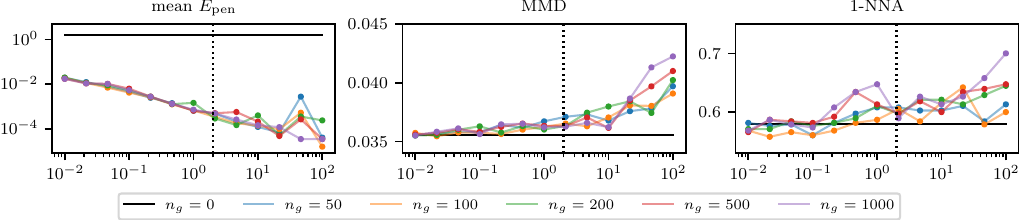}
    \caption{Hyperparameter sweep over guidance steps~$n_g$ and guidance weight $w_\text{pen}$ (x-axis) for generation with the penetration guidance loss only, without category conditioning. Black line: NAP baseline. We choose $n_g=500$ and $w_\text{pen,base}=2$ (dashed line). The $E_\text{pen}$ and MMD plots are identical to \Cref{fig:pen_sweep} in the main paper.}
  \label{fig:pen_sweep_wide}
\end{figure*}

\begin{figure*}[tb]
  \centering
  \includegraphics{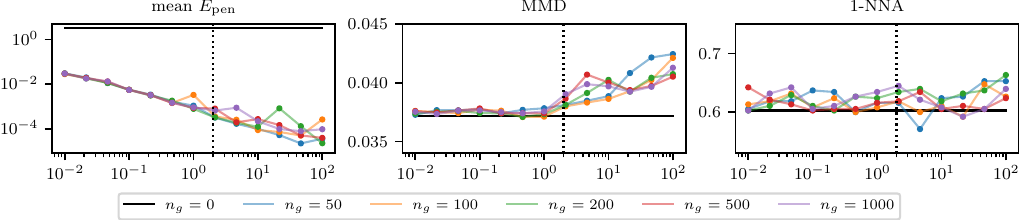}
    \caption{Hyperparameter sweep over guidance steps~$n_g$ and guidance weight $w_\text{pen}$ (x-axis) for generation with the penetration guidance loss only, with category conditioning. We choose $n_g=500$ and $w_\text{pen,base}=2$ (dashed line).}
  \label{fig:CAT_pen_sweep}
\end{figure*}

\begin{figure*}[tb]
  \centering
  \includegraphics{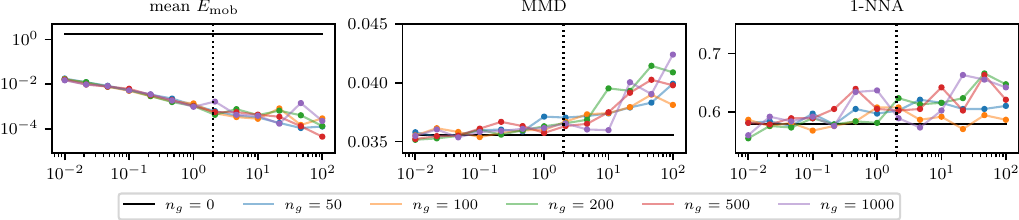}
    \caption{Hyperparameter sweep over guidance steps~$n_g$ and guidance weight $w_\text{mob}$ (x-axis) for generation with the mobility guidance loss only, without category conditioning. Black line: NAP baseline. We choose $n_g=500$ and $w_\text{mob,base}=2$ (dashed line).}
  \label{fig:dyn_sweep}
\end{figure*}

\begin{figure*}[tb]
  \centering
  \includegraphics{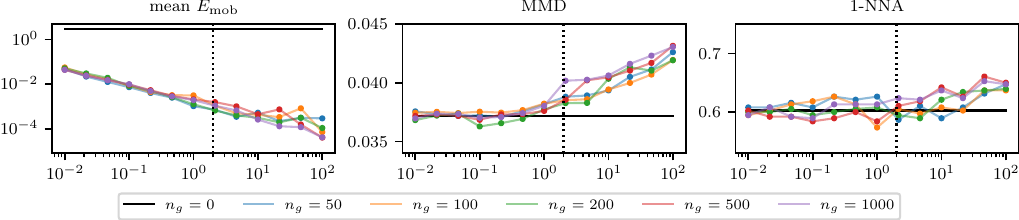}
    \caption{Hyperparameter sweep over guidance steps~$n_g$ and guidance weight $w_\text{pen}$ (x-axis) for generation with the mobility guidance loss only, with category conditioning. We choose $n_g=500$ and $w_\text{mob,base}=2$ (dashed line).}
  \label{fig:CAT_dyn_sweep}
\end{figure*}

\begin{figure*}[tb]
  \centering
  \includegraphics{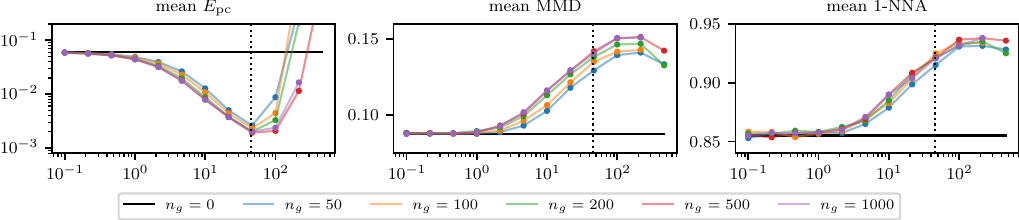}
  \caption{Hyperparameter sweep over guidance steps~$n_g$ and guidance weight $w_\text{pen}$ (x-axis) for generation with point cloud guidance loss only, without category conditioning. Black line: NAP baseline. We choose $n_g=500$ and $w_\text{pc,base}=45$ (dashed line). For high guidance weight and number of guidance steps, the generation may diverge to implausible results, which is the cause for missing points in this plot.}
  \label{fig:condper_sweep}
\end{figure*}

\begin{figure*}[tb]
  \centering
  \includegraphics{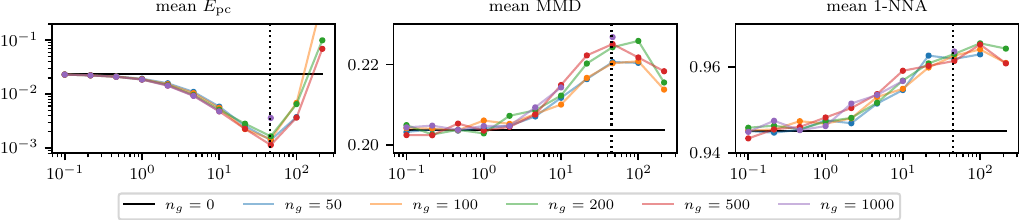}
  \caption{Hyperparameter sweep over guidance steps~$n_g$ and guidance weight $w_\text{pen}$ (x-axis) for generation with point cloud guidance loss only, with category conditioning. We choose $n_g=500$ and $w_\text{pc,base}=45$. For high guidance weight and number of guidance steps, the generation may diverge to implausible results, which is the cause for missing points in this plot.}
  \label{fig:CAT_condper_sweep}
\end{figure*}

\begin{figure*}[tb]
  \centering
  \includegraphics{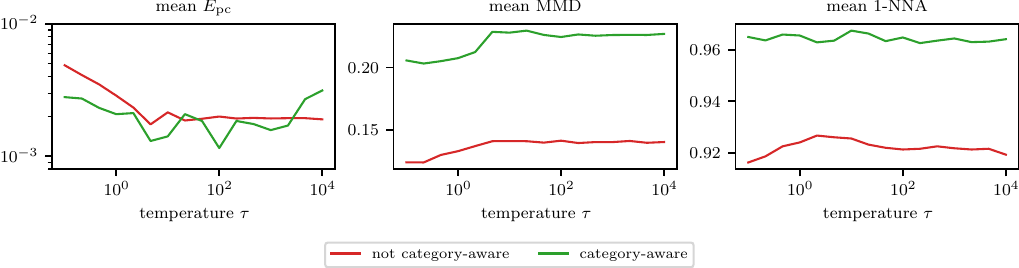}
  \caption{Hyperparameter sweep over inverse temperature parameter $\tau$ for generation with point cloud guidance loss only, with and without category conditioning. Other parameters are $n_g=500, w_{pc}=45$. We choose $\tau=1000$.}
  \label{fig:condper_temp}
\end{figure*}

\Cref{fig:pen_sweep_wide} to \Cref{fig:CAT_condper_sweep} show the effect of the two hyperparameters in the respective loss setting.
For our full method, we decide on the base hyperparameters $w_\text{pc,base} = 45, w_\text{pen,base}=2, w_\text{mob,base} = 2$.
\Cref{fig:condper_temp} shows the effect of the inverse temperature parameter $\tau$.
We considered values $\tau=10^2$ and $10^3$ as candidates due to simultaneously low mean $E_\text{pc}$, MMD and 1-NNA.
We conducted additional experiments on point clouds from the validation split for the pc and pc+pen+mob variants.
While the results are mixed, we found that $\tau=10^3$ performs better in 4 and worse in 3 value/metric combinations and similarly in the remaining ones, so we choose this value for further experiments.

\section{Mesh retrieval}
\begin{figure*}[tb]
    \centering
    \includegraphics[width=\linewidth]{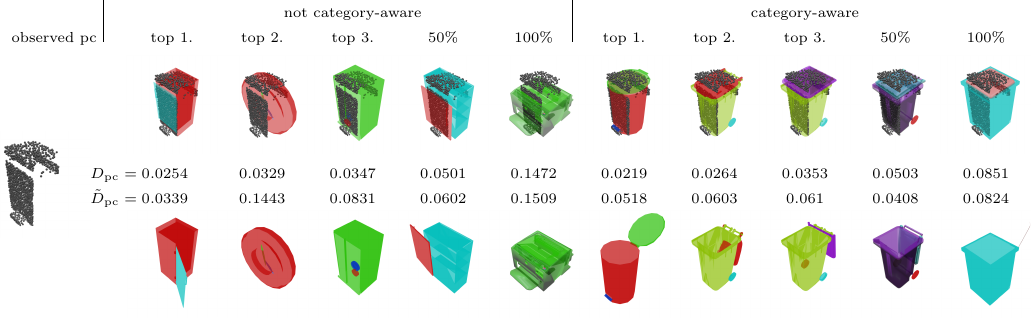}
    \noindent\rule{\textwidth}{0.5pt}
    \includegraphics[width=\linewidth]{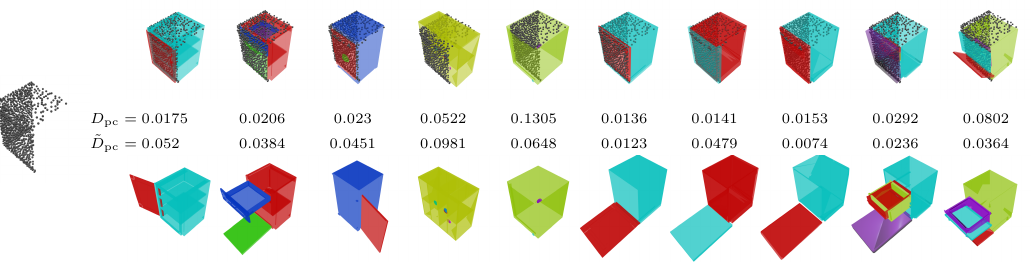}
    \noindent\rule{\textwidth}{0.5pt}
    \includegraphics[width=\linewidth]{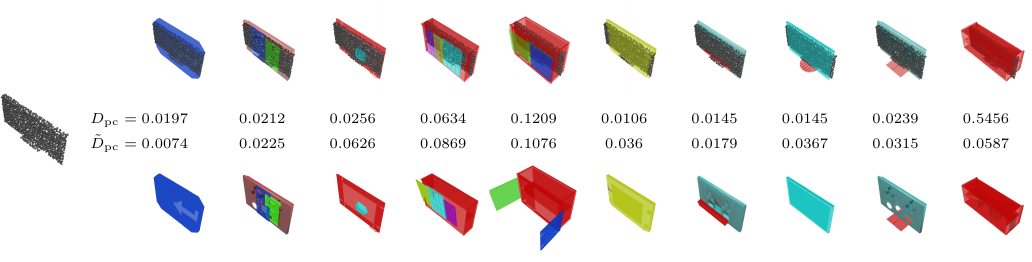}
    \noindent\rule{\textwidth}{0.5pt}
    \includegraphics[width=\linewidth]{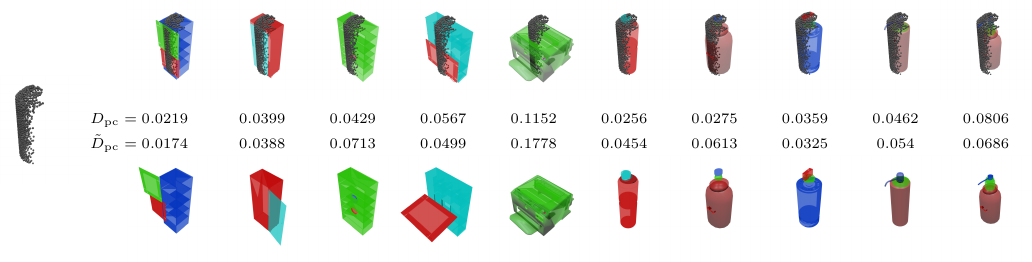}
    \caption{Shown are the same generated articulation graphs as in \Cref{fig:qualitative} in the main paper, but with mesh retrieval instead of SDF-based mesh extraction. We keep the sorting by $D_\text{pc}$ of the extracted objects and also provide those values for reference.
    In addition, we display the mean point distances to the retrieved meshes as $\tilde{D}_\text{pc}$. }
    \label{fig:qualitative_ret}
\end{figure*}

In \Cref{fig:qualitative_ret}, we show retrieved part meshes instead of meshes extracted from SDF latent codes and compare their mean distances to the observed point cloud.

\section{Run Times}
We provide the average run times of several ablations from \Cref{tab:main_results} in the main paper in \Cref{tab:runtime}.
The computations were all done on a cluster node with 16 CPU cores and an Nvidia A40 GPU.

\begin{table}[tb]
\centering
\footnotesize
\setlength{\tabcolsep}{3pt}
\begin{tabular}{llr}
\toprule
cat & variant & avg run time (m:ss)\\
\midrule
no & pc+pen+mob & 1:57\\
no & pc & 0:15\\
no & uncond & 0:13\\
\midrule
yes & pc+pen+mob & 1:31\\
yes & pc & 0:15\\
yes & uncond & 0:12\\
\bottomrule
\end{tabular}
\caption{Average run time per generated sample (minutes:seconds) for several variants from \Cref{tab:main_results} in the main paper.
Average taken over 150 samples (5 each for the 30 point clouds from the test set).
Adding the physical plausibility losses has a notable impact on run time.
The category-aware model is slightly faster than the category-unaware model.}
\label{tab:runtime}
\end{table}

\end{document}